\documentclass[journal]{IEEEtran}
\usepackage{cite}
\usepackage[pdftex]{graphicx}
\usepackage{subfigure}
\usepackage{amsmath}
\usepackage{amssymb}
\usepackage{algorithmic}
\usepackage{array}
\usepackage{multirow}

\usepackage{mathrsfs}
\usepackage{amsmath}

\begin{document}
%
\title{Body Joint guided 3D Deep Convolutional Descriptors for Action Recognition}
%
%
%

\author{
    Congqi Cao,
    Yifan Zhang,~\IEEEmembership{Member,~IEEE},
    Chunjie Zhang,~\IEEEmembership{Member,~IEEE},
    and Hanqing Lu,~\IEEEmembership{Senior Member,~IEEE}

    \thanks{\small{C. Cao, Y. Zhang (corresponding author) and H. Lu are with the National Laboratory of Pattern Recognition, Institute of Automation, Chinese Academy of Sciences, Beijing 100190, China
    (email: \{congqi.cao,yfzhang,luhq\}@nlpr.ia.ac.cn).
    C. Zhang is with the School of Computer and Control Engineering, University of Chinese Academy of Sciences
    (email: zhangcj@ucas.ac.cn). }}
}

\maketitle

\begin{abstract}

Three dimensional convolutional neural networks (3D CNNs) have been established as a powerful tool to simultaneously learn features from both spatial and temporal dimensions, which is suitable to be applied to video-based action recognition.
In this work, we propose not to directly use the activations of fully-connected layers of a 3D CNN as the video feature, but to use selective convolutional layer activations to form a discriminative descriptor for video.
It pools the feature on the convolutional layers under the guidance of body joint positions.
Two schemes of mapping body joints into convolutional feature maps for pooling are discussed.
The body joint positions can be obtained from any off-the-shelf skeleton estimation algorithm.
The helpfulness of the body joint guided feature pooling with inaccurate skeleton estimation is systematically evaluated.
To make it end-to-end and do not rely on any sophisticated body joint detection algorithm, we further propose a two-stream bilinear model which can learn the guidance from the body joints and capture the spatio-temporal features simultaneously.
In this model, the body joint guided feature pooling is conveniently formulated as a bilinear product operation.
Experimental results on three real-world datasets demonstrate the effectiveness of body joint guided pooling which achieves promising performance.

\end{abstract}


%
\IEEEpeerreviewmaketitle

\section{Introduction}

Recognizing the action performed in video is one of the most popular research fields in computer vision.
Different from images which only contain spatial information, videos are three dimensional (3D) spatio-temporal streams.
A lot of research focused on how to take both the appearance information and the motion information into account for video-based action recognition \cite{DBLP:journals/tcyb/ShaoZTL14,SimonyanZ2014_NIPS_TwoStream,cvpr/KarpathyTSLSF14_larege_sacle_video_classification,CNN_event_detection,C3D_2,DBLP:journals/tcyb/LiuSLL16}.

Much of the previous work on action recognition used hand-crafted features such as HOG \cite{cvpr/DalalT05_HOG}, HOF \cite{eccv/DalalTS06_HOF}, STIP \cite{cvpr/LaptevMSR08_STIP}, Dense Trajectories (DT) \cite{HengWang:CVPR2011_Dense_Trajectories} and Improved Dense Trajectories (IDT) \cite{wang:ICCV2013_iDT}.
DT has been shown to be an expressive video representation for action recognition. By taking camera motion into consideration, IDT improves the performance further.

Besides trajectories of dense points, human action can be represented by the trajectories of body joints which are more discriminative and compact. Human action consists of body poses and interactions with the environment. The position of body joints defines the appearance of pose, while the temporal evolution of body joints defines the motion.
Body joint coordinates can be obtained by skeleton estimation algorithms, \textit{e.g.} \cite{corr/TompsonJLB14_pose_estimation,cacm/ShottonSKFFBCM13_Real_Time_pose_recognition}.
Most of the existing skeleton-based action recognition methods model the appearance and the temporal dynamics of body joints with hand-crafted features, such as the relative locations between body joints, the angles between limbs and the angles between limbs and planes spanned by body parts \cite{composable_feature1}.
However, skeleton-based features are basically local features comprised of coordinates of body joints and their 2nd-order or high-order relations. Thus, they are not quite suitable for modeling and distinguishing actions with similar pose movement and human-object interactions, such as ``grab'' and ``deposit'' \cite{DBLP:conf/cvpr/DuWW15}. In addition, they heavily rely on the skeleton estimation algorithm. Inaccurate body joint detection cannot be well dealt with.



\begin{figure}[t]

  \centering
  \includegraphics[scale =1]{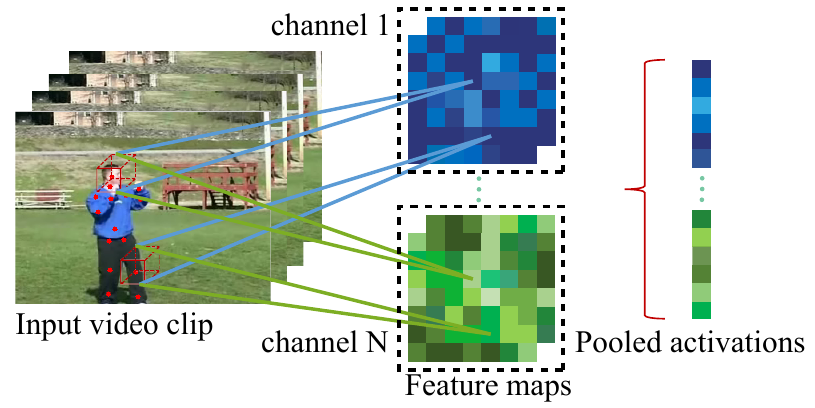}
  \caption{Illustration of body joint guided pooling in a 3D CNN. Different colors of feature maps represent different channels. Each channel of the feature maps is a 3D spatio-temporal cube. We pool the activations on 3D convolutional feature maps according to the positions of body joints. By aggregating the pooled activations of all the clips belonging to one video together, we obtain a descriptor of the video.}
  \label{fig:Framework}
\end{figure}

\begin{figure*}[t]
  \centering
  \includegraphics[width=\textwidth]{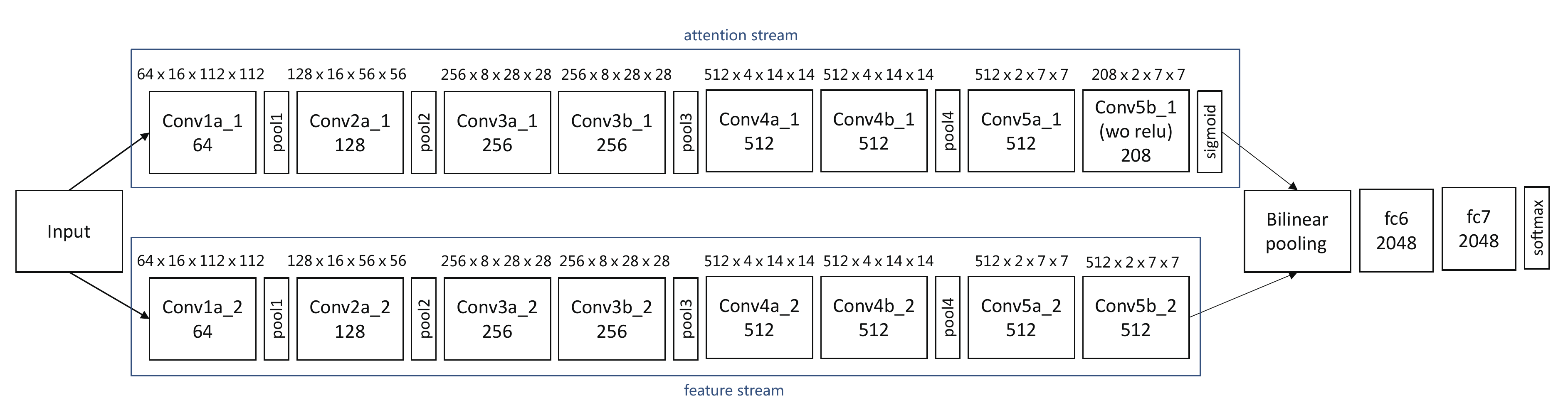}
  \caption{The block diagram of two-stream bilinear C3D. The attention stream is pre-trained to locate keypoints in 3D convolutional feature maps. It replaces the ReLU operation after the last convolutional layer of C3D with sigmoid ('wo' stands for 'without'). The feature stream inherits the convolutional structure of the original C3D to extract spatio-temporal features. The numbers inside the convolution blobs represent the number of channels, while the numbers above the blobs stands for the size of feature maps.}
  \label{block_diagram:two-stream}
\end{figure*}

CNNs have been proved to be effective in extracting local-to-global features.
Encouraged by the success of CNNs in image classification, recently much effort is spent on applying CNNs to video-based action recognition.
There are mainly two ways of applying CNNs to video data. One is using the 2D CNN architecture.
Directly applying image-based models to individual frames of videos can only characterize the visual appearance.
The two-stream CNN architecture \cite{SimonyanZ2014_NIPS_TwoStream} learns motion information by using an additional 2D CNN which takes the optical flow as input. The stacked optical flow frames are treated as different channels. After convolutional operation, the temporal dimension is collapsed completely. Therefore the two-stream CNN is less effective in characterizing long-range motion patterns among multiple frames. Furthermore, the computational complexity of optical flow is high.

The other way of adapting CNNs to video is using 3D CNNs with 3D convolution and 3D pooling layers \cite{cvpr/KarpathyTSLSF14_larege_sacle_video_classification,DBLP:journals/pami/JiXYY13,C3D_2}.
These layers take a volume as input and output a volume which can preserve the spatial and temporal information of the input.
Both the spatial information and the temporal information are abstracted layer by layer.
Tran \textit{et al.} \cite{C3D_2} found that the best architecture for 3D CNN is with small $3 \times 3 \times 3$ convolution kernels in all layers. A convolutional 3D network named as C3D was designed to extract features for videos.
The features used in \cite{C3D_2} were from fully-connected layers of C3D which achieved state-of-the-art performance on multiple video analysis tasks.
However, the weakness of fully-connected layers is the lack of spatio-temporal structure.
Compared with fully-connected layers, 3D convolutional layers preserve spatio-temporal grids. Different convolutional layers provide bottom-up hierarchical semantic abstraction.
If used appropriately, convolutional activations can be turned into powerful descriptors.
In image-based computer vision tasks, there have been explorations to utilize multiple convolutional layers for segmentation \cite{HariharanAGM15_Hypercolumns} and classification \cite{cross-layer-pooling}.
It is worth exploring how to utilize the spatio-temporal information in 3D convolutional layers to obtain discriminative features which combine different levels of abstraction.

In the preliminary version of our work \cite{IJCAI2016_JDD},
we have proposed an efficient way of pooling activations on 3D feature maps with the guidance of body joint positions to generate video descriptors as illustrated in Figure \ref{fig:Framework}.
The splitted video clips are fed into a 3D CNN for convolutional computation. The annotated or estimated body joints in the video frames are used to localize points in 3D feature maps. The activations at each corresponding point of body joint are pooled from every channel.
After aggregating the pooled activations of all the clips within the same video together, we obtain a descriptor of the video which is called joints-pooled 3D deep convolutional descriptor (JDD).
When mapping points from video into feature maps for pooling,
existing work \cite{/cvpr/Wang2015_TDD} used ratio scaling, in which only the sizes of the network's input and output are considered.
Different from this, we propose a novel method to map points by taking kernel sizes, stride values and padding sizes of CNN layers into account which is more accurate than ratio scaling.






In this paper, we extend our previous work \cite{IJCAI2016_JDD} as follows:
\textbf{1)} In \cite{IJCAI2016_JDD}, the body joint positions are obtained from either manual annotation or an off-the-shelf skeleton estimation algorithm. To make it do not rely on any sophisticated skeleton estimation algorithm, we propose a two-stream bilinear C3D model which can learn the guidance from the body joints and capture the spatio-temporal features simultaneously in this paper.
\textbf{2)} The body joint guided feature pooling is achieved by sampling (i.e. directly taking out the activations at the chosen positions corresponding to body joints) in \cite{IJCAI2016_JDD}. In this paper, the pooling process is formulated as a bilinear product operation in the proposed two-stream bilinear C3D model which is easy to be trained end to end.
\textbf{3)} We validate the effectiveness and good generalization capability of the two-stream bilinear C3D on three RGB datasets where not all the body joints are available.
\textbf{4)} An advanced version of aggregation method is introduced and analysed in this paper.
\textbf{5)} We systematically discuss the helpfulness of body joint guided pooling with inaccurate noisy skeletons under different accuracy rates.

The whole network of two-stream bilinear C3D is illustrated in Figure \ref{block_diagram:two-stream}.
The numbers inside the convolution blobs represent the number of filters, while the numbers above the blobs stands for the size of feature maps, in order of channel, length, height and width.
The first stream aims to predict keypoints in 3D feature maps which is pre-trained with the supervision of body joint positions. Since it functions as taking attention on discriminative regions automatically, we name it as an attention stream.
The second feature stream aims to capture spatio-temporal appearance and motion features which inherits the convolutional layers of the original C3D.
The two streams are multiplied with bilinear product.
It is end-to-end trainable with class label.


The main contributions of our work include:
\begin{itemize}
\item
We are the first to combine 3D CNNs and body joints to improve action recognition by using a novel method to map body joint positions in video frames to points in feature maps for pooling.
\item
We propose a two-stream bilinear model which can learn the guidance from body joints and capture the spatio-temporal features simultaneously.
\item
We formulate the pooling process in the proposed two-stream bilinear 3D CNN as a generalized bilinear product operation, making the model end-to-end trainable.
\end{itemize}

\section{Related Work}


Research in video-based action recognition is mainly driven by progress in image classification, where those approaches are adapted or extended to deal with image sequences.
Fine-grained recognition in images \cite{POOF,DBLP:conf/cvpr/DuanPCG12,Part-stacked_CNN,Fine-grained/ZhangSGD15} highlights the importance of spatial alignment which can increase the robustness toward image transforms.
Absorbing from these thoughts,
it is intuitive to align human poses with body joints to promote action recognition result.

There are two ways of extracting aligned features with CNN. One way is sampling multiple sub-images from the input image based on keypoints (such as the body parts of birds) and using one CNN to extract features for each sub-image \cite{Multi-scale_Orderless_Pooling,Encode_Local_Features,PANDA}; encoding is employed to aggregate regional features, which usually are fully-connected layer activations, to image-level representation. Another option is only taking the whole image as input and pool convolutional activations with the guidance of keypoints on feature maps \cite{PosePoolingKernelsZhangEtalCVPR12,Par-based_RCNNs/ZhangDGD14,PANDA,cross-layer-pooling}. Since the latter method only needs to run a CNN once for an image, it reduces the computational cost compared with the former one which needs to run CNN forward computation multiple times.

To extract representation of video for action recognition, the pose-based CNN descriptor P-CNN \cite{/corr/CheronLS15/PCNN} cropped RGB and optical flow images into multiple part patches (\textit{e.g.} right hand and left hand) as the inputs of a two-stream CNN. This belongs to the first alignment method mentioned above.
Different from this, we take advantage of the abundant information in 3D convolutional feature maps by body joint guided pooling.
Our approach is more efficient in computation than P-CNN that needed to compute not only body joint positions but also optical flow images and uses multiple inputs with two deep 2D CNNs. Furthermore, we do not need to compute activations of fully-connected layers.

Trajectory-pooled deep-convolutional descriptors (TDD) \cite{/cvpr/Wang2015_TDD} falls within the second alignment scheme. Wang \textit{et al.} utilized two-stream CNN to learn convolutional feature maps, and conduct trajectory-constrained pooling to aggregate these convolutional features into video descriptors.
The main differences between TDD and JDD are:
Firstly, TDD used 2D CNNs, while we adopt 3D CNNs. 3D CNNs are more suitable for spatio-temporal feature learning compared with 2D CNNs owning to 3D convolution and 3D pooling operations.
Secondly, TDD used dense trajectory points to pool the feature maps, while we use body joints. For human action recognition, body joints are more discriminative and compact compared with dense trajectory points.
Thirdly, TDD used ratio scaling to map trajectory points into feature maps. We compute the corresponding points of body joints by taking the kernel sizes, strides and paddings of CNN layers into consideration, which is more accurate. In addition, we do not need to compute optical flow while TDD needed.

Without directly using the annotation of keypoints, Liu \textit{et al.} \cite{cross-layer-pooling} adopted to pool one convolutional layer with the guidance of the successive convolutional layer for fine-grained image recognition.
Bilinear CNN \cite{bilinear_CNN} can be seen as a generalization of cross-convolutional-layer pooling \cite{cross-layer-pooling} which multiplied the outputs of two separate CNNs using outer product at each location and pooled to obtain an image descriptor. By using two CNNs, the gradient computation of bilinear CNN is simplified for domain specific fine-tuning  compared with cross-convolutional-layer pooling.
\cite{Part-stacked_CNN} and \cite{Fine-grained/ZhangSGD15} took an extra step to train one of the two CNNs in bilinear model explicitly with part annotations for keypoints prediction. All these works aim at fine-grained image classification.


It should be noted that the recently popular concept of $soft$ $attention$ $based$ $model$ \cite{AR_visual_attention,AR_visual_attention_2} can also fit into the framework of keypoint pooling which used Recurrent Neural Networks (RNN) to learn attention regions of the video frames sequentially. In these work, one group of pooling weights for the current frame was learned at each time.
While we utilize 3D CNN to learn multiple groups of pooling weights which correspond to  multiple spatio-temporal attention regions for video clip as an extension of our previous work in this paper.

\section{Joints-pooled 3D Deep Convolutional Descriptors}

In this section, we give an introduction to the proposed joints-pooled 3D deep convolutional descriptors (JDD). Firstly, we review the 3D architecture proposed in \cite{C3D_2}. Then we introduce two schemes of mapping body joints to points in feature maps for pooling. Finally we describe two methods of feature aggregation.

\subsection{3D Convolutional Networks Revisited}

We use the C3D model proposed by Tran \textit{et al.} \cite{C3D_2} which is trained on Sports-1M dataset to compute 3D convolutional feature maps.

Using shorthand notation, the full architecture of C3D is $conv1a(64)-pool1-conv2a(128)-pool2-conv3a(256)-conv3b(256)-pool3-conv4a(512)-conv4b(512)-pool4-
conv5a(512)-conv5b(512)-pool5-fc6(4096)-fc7(4096)-softmax$, where the number in parenthesis indicates the number of convolutional filters.
There is a ReLU layer after each convolutional layer which is not listed.
All 3D convolution kernels are $3 \times 3 \times 3$ (in the manner of $d \times k \times k$, where $d$ is temporal depth and $k$ is spatial size) with stride $1$ and padding $1$ in both spatial and temporal dimensions.
All pooling kernels are $2 \times 2 \times 2$, except for $pool1$ which is $1 \times 2 \times 2$ with the intention of not to merge the temporal signal too early.
C3D takes a clip of 16 frames as input. It resizes the input frames to $171 \times 128$px (width $\times$ height), then crops to $112 \times 112$. More details and explanations can be found in \cite{C3D_2}.



\subsection{Mapping Schemes}  \label{section:coordinate mapping}

For JDD, we compare two schemes of mapping body joints to points in 3D convolutional feature maps.
One straightforward way is using the ratio of the network's output to its input in spatial and temporal dimensions to scale the body joint coordinates from the original video frame into feature maps as shown in Equation \ref{equation:ratio scaling}, which is named as \textbf{Ratio Scaling}.

\begin{equation}
(x^i_c,y^i_c,t^i_c) = (\overline{(r_x^i \cdot {x_v})}, \overline{(r_y^i \cdot {y_v})}, \overline{(r_t^i \cdot {t_v})})
\label{equation:ratio scaling}
\end{equation}
where $\overline{()}$ is the rounding operation
and $(x^i_c,y^i_c,t^i_c)$ is the point coordinate in the $i$th 3D convolutional feature maps corresponding to $(x_v,y_v,t_v)$ which is the body joint coordinate in the original video clip. $(r_x^i,r_y^i,r_t^i)$ is the size ratio of the $i$th convolutional feature maps to the video clip in spatial and temporal dimensions.

A more precise way to compute the accurate coordinate of the point corresponding to body joint, is taking the kernel size, stride and padding of each layer into account. We call it as \textbf{Coordinate Mapping}.

The mapping relationship of points can be computed layer by layer.
Let $p_{i}$ be a point in the $i$th layer.
$(x_i,y_i,t_i)$ is the coordinate of $p_{i}$.
Given $p_i$, the corresponding point $p_{i+1}$ can be computed by mapping $p_i$ to the $(i+1)$th layer.

For the convolutional layers and pooling layers, the coordinate mapping from the $i$th layer to the $(i+1)$th layer is formulated as follow:

\begin{equation}
x{}_{i + 1} = \frac{1}{{s_i^x}}({x_i} + padding_i^x - \frac{{k_i^x - 1}}{2})
\label{ConvPool_CoordinateMapping}
\end{equation}
where $s^x_i, k^x_i, paddin{g^x_i}$ are the $x$-axis component of stride, kernel size and padding of the $i$th layer respectively.
The equation also applies to $y$ and $t$ dimensions.
Similar to Equation \ref{equation:ratio scaling}, the coordinate in the left of equals sign should be the value after rounding. For clarity, we omit the symbol of rounding in the equation.

For ReLU layers, since the operations do not change the size of the feature maps, the coordinate mapping relationship is unchanged between layers, which is formulated as follow:

\begin{equation}
(x_{i+1},y_{i+1},t_{i+1}) = (x_{i},y_{i},t_{i})
\label{Neuron_CoodinateMapping}
\end{equation}



We need to take all the preceding layers into consideration to compute the coordinate mapping relationship between feature maps and video clip.
In C3D, all the convolutional layers use $3 \times 3 \times 3$ kernels sliding in spatial and temporal dimensions with stride 1 and padding 1. The strides of pooling layers are 2 in both spatial and temporal dimensions and there is no padding when pooling. The kernel sizes of pooling layers are $2 \times 2 \times 2$ except for $pool1$ which uses $1 \times 2 \times 2$ kernels without merging the signal in temporal dimension.
After bringing the values of C3D kernel sizes, strides and paddings into Equation \ref{ConvPool_CoordinateMapping} and Equation \ref{Neuron_CoodinateMapping} recurrently,
we can obtain the relationship between point coordinates in the $i$th convolutional feature maps and body joint positions in the input video clip, which is formulated as follows:

\begin{equation}
(x^i_c,y^i_c) = \frac{1}{{{2^{i - 1}}}} \cdot ({x_v} - \frac{{{2^{i - 1}} - 1}}{2},{y_v} - \frac{{{2^{i - 1}} - 1}}{2})
\label{equation:coordinate mapping spatial dim}
\end{equation}
\begin{equation}
t^i_c = \frac{1}{{{2^{i - 2}}}}({t_v} - \frac{{{2^{i - 2}} - 1}}{2})
\label{equation:coordinate mapping temporal dim}
\end{equation}

The coordinates in the left of equals sign should be the values after rounding. We omit the operation of rounding in the equations for clarity.
The expression in temporal dimension is a little different from that in spatial dimension because the input is downsampled in the spatial dimension one more time than that in the temporal dimension.

\subsection{Aggregation Methods}     \label{Section:aggregation_methods}

To recognize the action performed in a video sequence, since the video is splitted into clips as the input of C3D, we need to aggregate the extracted features of clips over time to form a video descriptor for classification.

By employing body joints in video clips to localize points in 3D feature maps, we can determine the positions to pool.
The pooled representation corresponding to each body joint in a frame of a video clip is a $C$ dimensional feature vector, where $C$ is the number of feature map channels. We use $f_k^{i,t}$ to represent the $C$ dimensional feature vector pooled with the guidance of the $i$th body joint at the $t$th frame of the $k$th clip.

There are two ways to aggregate the pooled feature vectors in all the clips within a video sequence to a video descriptor.
One way is to follow the setting of C3D \cite{C3D_2}. It is straightforward to concatenate all the pooled feature vectors belonging to one clip as a representation of the clip. It is a $C \times N \times L$ dimensional feature formulated as $f_k = {[f_k^{1,1}, f_k^{2,1}, ..., f_k^{N,1}, f_k^{1,2}, f_k^{2,2}, ..., f_k^{N,2}, ..., f_k^{N,L}]}$, where $N$ is the number of body joints in each frame and $L$ is the length of the video clip. Then average pooling and L2 normalization are used to combine $K$ clip representations ${\{f_1, f_2, ..., f_K\}}$ into a video descriptor, where $K$ is the number of clips within the video sequence.
The dimension of JDD aggregated in this way is $C \times N \times L$. This is the default aggregation method for JDD. Unless otherwise specified, we use this aggregation method for a fair comparison with the original C3D features.

The other way of aggregation is to concatenate the pooled feature vectors corresponding to the body joints in one frame as a $C \times N$ dimensional representation which is formulated as $f_k^t = {[f_k^{1,t}, f_k^{2,t}, ..., f_k^{N,t}]}$.
Then one clip is characterized by $L$ representations $\{f_k^{1}, f_k^{2}, ..., f_k^{L}\}$ within the same clip.
Max+min pooling is used to aggregate these representations into a clip descriptor, where max+min pooling means selecting the maximum value and the minimum value of each feature dimension.
The whole video contains $K$ clips in total.
Finally max+min pooling and L2 normalization are used to combine the $K$ clip descriptors into a video descriptor.
In this way, the dimension of JDD is $4 \times C \times N$ which is independent of the length of the video clip since the activations are pooled along time not only between clips but also within clip. The dimension of JDD is quadruple that of $f_k^t$ because max+min pooling is used twice. We choose max+min pooling for its good performance compared with average pooling and max pooling which has been verified in \cite{/corr/CheronLS15/PCNN}. This is the advanced version of aggregation for JDD.



In addition, the JDDs obtained from different convolutional layers can be fused together to improve the ability of representation due to their complementarity. JDDs can also be combined with other features or models.

\section{Two-stream bilinear C3D}

In order to get rid of the dependence of complicated skeleton estimation algorithms, a framework consisting of two C3D streams multiplied with bilinear product is designed to take attention on keypoints, extract features and generate the pooled descriptors for video clips jointly which can be trained end to end using class label.
In this section, firstly, we explain how to compute JDD with bilinear product.
Then, we detail the network we apply to automatically predict spatio-temporal keypoints in 3D convolutional feature maps with the guidance of body joints.
Finally, we introduce the forward and backward computations of a general bilinear formulation used in our model.

\subsection{JDD by Bilinear Product}  \label{section:JDD_with_bilinear_product}


The original body joint guided feature pooling in JDD is realized by selecting the activations at the corresponding points of body joints on convolutional feature maps, which is equivalent to assigning hard weights to the activations.
Specifically, the activations at the corresponding points of body joints are assigned with weight 1, while the activations not at the corresponding points of body joints are assigned with weight 0.

Given a video clip, we can generate a heat map with the same spatio-temporal size of the convolutional feature maps to be pooled for each body joint at each frame.
We represent the size with $l \times h \times w$, in order of length, height and width.
In the heat map, the value at the corresponding point of the body joint is coded as $1$, while the others are coded as $0$.
Then the process of pooling on one feature map guided by the heat map of one body joint can be formulated as a pixel-wise product between the 3D feature map and the 3D heat map, followed by a summation over all the pixels.
If we flatten all the activations on the 3D feature map and the activations on the 3D heat map into a vector respectively, the above operations can be considered as an inner product between the two vectors.

For JDD, there are $M$ channels of heat maps in total, where $M=N \times L$ with $N$ representing the number of body joints in each frame and $L$ standing for the length of video clip.
With heat maps, the computation of JDD can be achieved through bilinear product.
Firstly, resize the heat maps with the size of $M \times l \times h \times w$ to a 2D matrix $\mathcal{A}$ with $M$ rows and $l \times h \times w$ columns. Similarly, resize the $C \times l \times h \times w$ feature maps derived from the original C3D to a 2D matrix $\mathcal{B}$ with $C$ rows and $l \times h \times w$ columns.
Then the bilinear product can be formulated as:

\begin{equation}
\mathcal{P} = \mathcal{A}{\mathcal{B}^T}
\label{equation:bilinear_product}
\end{equation}
where $\mathcal{B}^T$ is the transposition of $\mathcal{B}$. $\mathcal{P}$ is a matrix with the size of $M \times C$.

After concatenating all the values in the product matrix $\mathcal{P}$ as a long vector, the identical feature to JDD for video clip is obtained. The feature can also be formulated as $L$ representations with $N \times C$ dimensions as the same with JDD.

Note that the heat maps can be viewed as $M$ groups of weights.
With heat maps, it is convenient to generalize hard pooling to soft pooling by assigning convolutional activations soft weights within the range of [0,1].
Not only JDD, any keypoints pooling or attention pooling methods can be computed with bilinear product as described above.


\begin{figure}[t]
  \centering
  \includegraphics[width=0.48\textwidth]{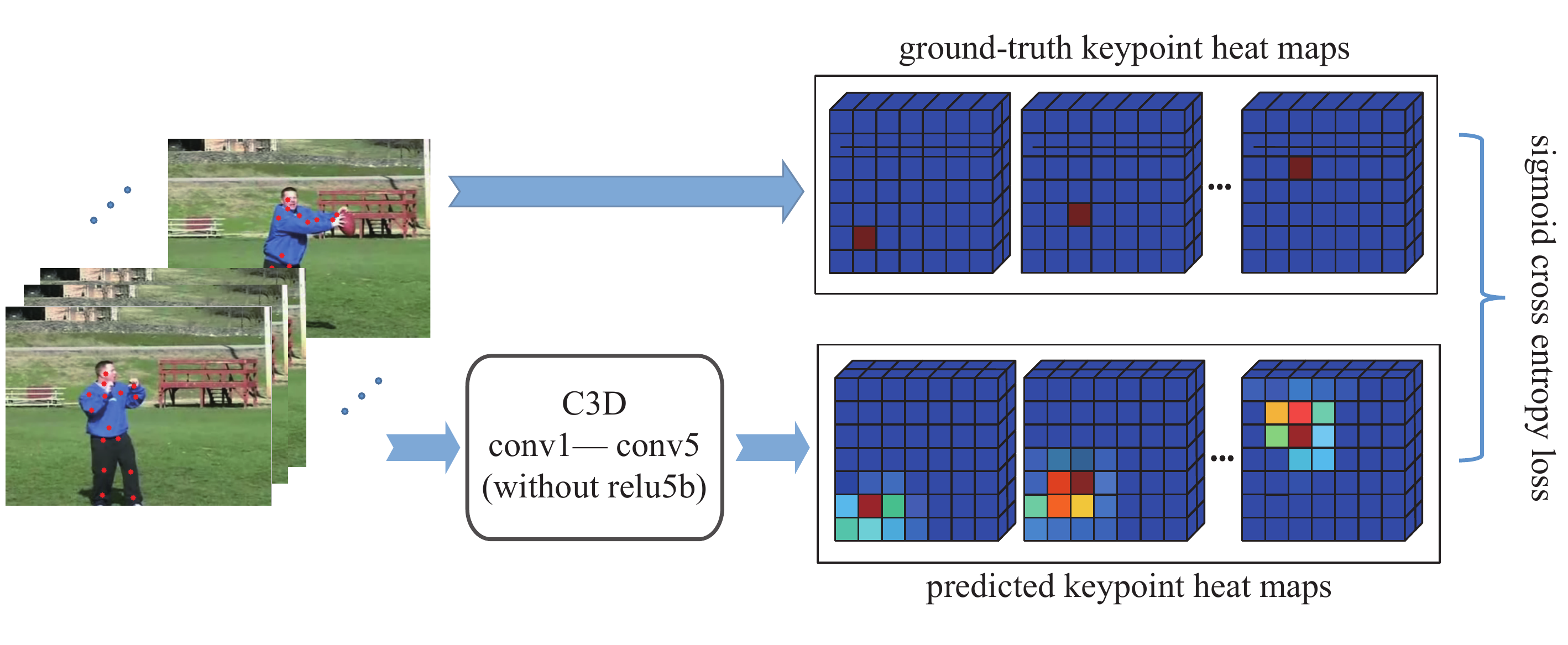}
  \caption{The attention model to predict spatial-temporal keypoints in 3D convolutional feature maps. It is pre-trained with the supervision of body joint positions. A 3D CNN is used to regress the ground-truth heat maps of keypoints corresponding to the annotated body joints.}
  \label{fig:3D_attention_model}
\end{figure}

\subsection{3D attention model}   \label{section:3D_attention_model}

For body joint guided pooling, we can use manually annotated or automatically detected body joints.
However, not all the datasets and practical situations can offer annotated body joints, and the computation complexity of skeleton estimation algorithms are high in general.
Actually, we only need to locate the keypoints in the downsampled feature maps instead of to estimate the precise body joint positions in the original video frames.
Therefore, we propose to use a 3D CNN to take attention on discriminative parts in feature maps automatically which is pre-trained with the guidance of body joint positions.

A C3D without the ReLU layer after $conv5b$ and fully-connected layers is used to regress the heat maps of the corresponding points described in Section \ref{section:JDD_with_bilinear_product}, separately for each body joint in an input video clip.
We use a pixel-wise sigmoid cross entropy loss for training:

\begin{eqnarray}
\ell = \frac{1}{M}\sum\limits_{m = 1}^M {\sum\limits_{k = 1}^l {\sum\limits_{j = 1}^h {\sum\limits_{i = 1}^w {(p_{ijk}^m\log \hat p_{ijk}^m} } } } \nonumber \\
+ (1 - p_{ijk}^m)\log (1 - \hat p_{ijk}^m))
\label{equation:sigmoid_cross_entropy_loss}
\end{eqnarray}
where $p_{ijk}^m$ is the ground-truth heat map value of the $m$th channel at location $(i,j,k)$ and $\hat p_{ijk}^m$ is the sigmoid value of $conv5b$'s output at location $(i,j,k)$ of the $m$th channel. $i,j,k$ are the index in width, height, length respectively.

The outputs of the 3D attention model are soft weight heat maps corresponding to each body joint.
By using the heat maps to pool the original C3D's convolutional feature maps with bilinear product, we can obtain the descriptors of video clips independent of skeleton estimation algorithms.
This architecture is indeed a two-stream bilinear C3D model.

\subsection{End-to-End training of two-stream bilinear C3D}

We propose a two-stream bilinear C3D model to learn the guidance from the body joints and capture the spatio-temporal features automatically.
The two streams (i.e. the attention stream which inherits the parameters of the pre-trained 3D attention model and the feature stream which inherits the original C3D's convolutional layers) are combined with bilinear product.
The whole network can be trained end-to-end with class label. In this section, we will introduce the forward and backward computations of a general bilinear product used in our model.

The forward computation of a general form of bilinear product is:

\begin{equation}
\mathcal{P} = \mathcal{AW}{\mathcal{B}^T}
\label{equation:bilinear_product_in_general}
\end{equation}
where $\mathcal{P} \in {\mathbb{R}^{M \times C}}, \mathcal{A} \in {\mathbb{R}^{M \times {K_1}}}, \mathcal{W} \in {\mathbb{R}^{{K_1} \times {K_2}}}, \mathcal{B} \in {\mathbb{R}^{C \times {K_2}}}$. $\mathcal{W}$ is a matrix of parameters to be learned by the network. Compared with Equation \ref{equation:bilinear_product},  $\mathcal{A}$ and $\mathcal{B}$ need not to have the same number of columns in this formulation. Furthermore, with $\mathcal{W}$, more statistical characteristics between $\mathcal{A}$ and $\mathcal{B}$ can be learned.

We use the general form of bilinear product in our proposed two-stream bilinear C3D model. Figure \ref{block_diagram:two-stream} illustrates the architecture of the network which includes attention, feature extraction, keypoints pooling and classification in a unified framework.

The whole network is end-to-end trainable with softmax loss supervised by class label.
The gradients of the bilinear layer can be computed with back propagation as follow:

\begin{eqnarray}
\mathcal{\frac{{\partial \ell }}{{\partial A}}} &=& \mathcal{\frac{{\partial \ell }}{{\partial P}}B}{\mathcal{W}^T} \nonumber \\
\mathcal{\frac{{\partial \ell }}{{\partial B}}} &=& {\mathcal{(\frac{{\partial \ell }}{{\partial P}})}^T}\mathcal{AW} \nonumber \\
\mathcal{\frac{{\partial \ell }}{{\partial W}}} &=& {\mathcal{A}^T}\mathcal{\frac{{\partial \ell }}{{\partial P}}B}
\label{equation:bilinear_product_backprop}
\end{eqnarray}
where $\mathcal{\frac{{\partial \ell }}{{\partial A}}}$ is the back propagated gradient of the loss function ${\ell}$ wrto. $\mathcal{A}$.

\section{Experiments} \label{section:experiments}

\newcommand{\tabincell}[2]{\begin{tabular}{@{}#1@{}}#2\end{tabular}}

\begin{table*}[t]
\begin{center}
\caption{Recognition accuracy of baselines and JDDs with different configurations on subJHMDB dataset.}
\label{table:exploration_on_subJHMDB}
\begin{tabular}{|c|c c c c c|}
  \hline
   & \tabincell{c}{Concatenate\\all the activations} & \tabincell{c}{JDD\\Ratio Scaling\\$(1 \times 1 \times 1)$} & \tabincell{c}{JDD\\Coordinate Mapping\\$(1 \times 1 \times 1)$} & \tabincell{c}{JDD\\Ratio Scaling\\$(3 \times 3 \times 3)$} & \tabincell{c}{JDD\\Coordinate Mapping\\$(3 \times 3 \times 3)$}
  \\
  \hline
  joint coordinates    & {0.5480} & -      & -      & -      & -               \\
  fc7    & {0.6892} & -      & -      & -      & -               \\
  fc6    & {0.6996} & -      & -      & -      & -               \\
  conv5b & {0.6713} & {0.7724} & {\textbf{0.8186}} & {0.7843} & {0.8007} \\
  conv5a & {0.5952} & {0.7296} & {0.7563} & {0.7273} & {0.7343}               \\
  conv4b & {0.5057} & {0.7349} & {0.7251} & {0.7507} & {0.7773}          \\
  conv3b & {0.3943} & {0.6696} & {0.6419} & {0.6730} & {0.6733}          \\
  \hline
\end{tabular}
\end{center}
\end{table*}

In this section, we firstly make a brief introduction to the datasets and the experimental settings we use.
Then, we compare the two mapping schemes with exploratory experiments.
We also analyse the performance of JDD with feature fusion.
Next, we test the robustness of JDD with estimated body joints as well as by adding random white Gaussian noise to the ground-truth body joint positions.
Besides these, the experimental results of two-stream bilinear C3D and other pooling models are reported.
Finally, we evaluate JDD and two-stream bilinear C3D on public datasets and give a comparison with the state-of-the-art results.


\subsection{Datasets}  \label{section:dataset}
We evaluate our method on three public action datasets: subJHMDB \cite{Jhuang_2013_ICCV_subJHMDB_dataset}, Penn Action \cite{iccv/ZhangZD13_Penn_Action_dataset} and UCF101 dataset \cite{UCF101_dataset}.
The scales of the three datasets increase successively.
These datasets all cover actions performed indoor and outdoor.
The first two datasets are provided with annotated body joints, while the last one is not.


\textbf{subJHMDB dataset} is a dataset with full body inside the frame, containing 316 videos with 12 action categories. subJHMDB provides action labels for each video and the annotation of 15 body joints for each frame. We use the 3-fold cross validation setting provided by the dataset for experiments. The dataset is collected from movies and Internet. The lengths of frames in videos range from 16 to 40.

\textbf{Penn Action dataset} contains 2326 video sequences of 15 action classes. 13 body joints are annotated for each frame, even though there are videos in which not all the body joints are inside the frame or visible. We use the 50/50 trainning/testing split provided by the dataset to evaluate the results on it.
For other dataset which are without annotated body joints or too small to train an attention model, we split Penn dataset to 90/10 training/testing split randomly to train an attention model used for transferring.
The videos are obtained from various online video repositories. The lengths of videos are from 18 to 663 frames.

\textbf{UCF101 dataset} consists of 13,320 videos belonging to 101 human action categories. There is no annotation of body joints and it is time-consuming to estimate the positions of body joints for this relatively large dataset. The attention model trained on Penn Action dataset is transferred to it for keypoint pooling. We use the three split setting provided with this dataset. The video lengths are from 29 to 1776 frames.


\subsection{Implementation details}   \label{section:implementation_details}

We split the videos into clips as the input of 3D CNN models. The videos of Penn Action and UCF101 datasets are splitted into 16-frame clips with 8-frame overlapping between two consecutive clips.
For subJHMDB, due to its short length (34 frames in average), we sample the first, the middle and the last 16 frames of each video to generate three clips, where the consecutive two clips may have several frames overlapped.

When extracting JDD, each clip is input to C3D. We pool the activations of a particular 3D convolutional layer based on body joint positions.
The pooled activations of all the clips belonging to one video are aggregated to be a descriptor of the video based on the methods introduced in Section \ref{Section:aggregation_methods}.
Linear SVM \cite{SVM_Liblinear} is used to classify the video descriptors.
Since subJHMDB dataset and Penn Action dataset are provided with annotated body joints, comparison between JDDs based on the ground truth and estimated body joints can be made on these datasets.
The skeleton estimation algorithm \cite{/cvpr/YangR11/subJHMDB/estimated_joints} based on RGB images is used to predict the positions of body joints.
The computational complexity of \cite{/cvpr/YangR11/subJHMDB/estimated_joints} is $O(L \times T^2)$ with L locations and T mixture component types for each body part.
The runtime is approximately 10 seconds for 15 body joints without GPU acceleration on Intel i7-3770 CPU @3.4GHz.
Note that we do not finetune the original C3D for feature extraction on subJHMDB and Penn Action datasets.

As for two-stream bilinear C3D, the attention stream is pre-trained with the body joints of Penn Action dataset. And the whole network is finetuned with class label on Penn Action dataset. To test the generalization capability of the 3D attention model trained on Penn dataset, the attention model is transferred to subJHMDB and UCF101 datasets for keypoints prediction and bilinear pooling.

\subsection{Analysis of JDD and baselines}

The body joint coordinates and C3D features are used as baselines.
We compare JDD with these features and evaluate JDD with different pooling settings.
The results on subJHMDB dataset are listed in Table \ref{table:exploration_on_subJHMDB}.
We adjust the parameters of linear SVM for each kind of feature by cross validation.

The recognition accuracies of using body joint coordinates as feature and C3D features are listed in the first column of Table \ref{table:exploration_on_subJHMDB}. We can see that directly using the coordinates of body joints as feature performs not well. With the increase of layers, the C3D features, which are obtained by concatenating all the activations of a specific layer as a long vector, are more discriminative since they achieve better results. This implicitly indicates that deep architectures learn multiple levels of abstraction layer by layer. The recognition accuracy of $fc7$ is a little inferior to that of $fc6$. It is probably because we do not finetune the original C3D on subJHMDB dataset that the second fully-connected layer is more suitable for the classification of the pre-trained dataset.

For JDD, we show the experiments on pooling at different 3D convolutional layers with different body joint mapping schemes.
We also test JDD with pooling one activation at the corresponding point in feature maps and pooling a $3 \times 3 \times 3$ cube around the corresponding point, where corresponding point refers to the point in feature maps corresponding to body joint in the original video frame.
From Table \ref{table:exploration_on_subJHMDB}, we can see that, compared with C3D features, our proposed JDDs have superior performance which demonstrates the effectiveness of body joint guided pooling.
Generally, JDDs pooled from the higher layers encapsulate more discriminative information for recognition. Pooling a cube around the corresponding points is usually better than pooling one activation because the former takes more surrounding information into account, except for $conv5b$ and $conv5a$. It is probably because the spatial and temporal size of the feature maps in these layers is small, thus a cube around the corresponding point encloses too much background information which impairs the performance.
At shallow layers, JDDs with Ratio Scaling and JDDs with Coordinate Mapping are close in performance. As the layer goes deeper, the performance of JDDs with Coordinate Mapping is much better than that with Ratio Scaling. It is probably because the difference between the coordinates of corresponding points obtained by the two mapping schemes becomes more significant on the deeper layer.
The best result is obtained by JDD of $conv5b$ with Coordinate Mapping.
For other layers, the performance of JDDs obtained by pooling a cube around the corresponding point with Coordinate Mapping is also better than that with Ratio Scaling. This verifies that Coordinate Mapping is more appropriate than Scaling Ratio for computing the coordinates of corresponding points.
We use Coordinate Mapping and  $1 \times 1 \times 1$ pooling for $conv5b$, $3 \times 3 \times 3$ pooling for other layers in the rest experiments.


\begin{table}[t]
\begin{center}
\caption{Recognition accuracy of fusing JDDs from multiple layers together on subJHMDB dataset.}
\label{table:fusion}
\begin{tabular}{|c|c|}
  \hline
  Fusion Layers & Accuracy
  \\
  \hline
  JDD (conv5b+fc6)    & 0.825 \\
  JDD (conv5b+conv4b) & \textbf{0.833} \\
  JDD (conv5b+conv3b) & 0.830 \\
  \hline
\end{tabular}
\end{center}
\end{table}

Additionally, we try to fuse JDDs from different convolutional layers together to see if they can compensate each other.
Table \ref{table:fusion} represents the results of different combinations using late fusion with the scores of SVM on subJHMDB.
Fusing JDDs of different layers indeed improves the recognition results, which indicates that the features are complementary.
The combination of JDDs from $conv5b$ and $conv4b$ improves the performance mostly.

\subsection{Robustness analysis of JDD}    \label{section:robustness_analysis}

To evaluate the influence of body joint detection's precision to JDD, we generate JDD based on the estimated body joints computed by \cite{/cvpr/YangR11/subJHMDB/estimated_joints}.
For subJHMDB dataset, the per joint L1 distance error between the ground-truth body joint positions and the estimated body joint positions is (22.93, 17.57) pixels in width and height.
The average ratio of L1 distance error to frame size is (0.072, 0.073).
The accuracies of JDD based on ground truth and estimated body joints are listed in Table \ref{table:estimated_vs_groundtruth}. The drop of accuracy is also reported.
We compare JDD with P-CNN \cite{/corr/CheronLS15/PCNN}, Pose \cite{Jhuang_2013_ICCV_subJHMDB_dataset} and HLPF \cite{/corr/CheronLS15/PCNN}.
P-CNN is a pose-based CNN descriptor which used positions of body joints to crop RGB and optical flow images into patches. These patches are used as multiple inputs to feed into two 2D CNNs. The activations of fully-connected layers and their temporal differences were aggregated with max+min pooling to describe the video.
Pose and HLPF are hand-crafted high-level pose features.
They share the same idea to design features:
The distances and inner angles between the normalized body joint positions were used as static features;
dynamic features were obtained from trajectories of body joints;
temporal differences of some static features were also combined.
Compared with Pose, HLPF used head instead of torso as the center to compute relative positions; HLPF also converted angles from degrees to radians and L2 normalized the features.
In \cite{Jhuang_2013_ICCV_subJHMDB_dataset}, the authors verified that the high-level pose features outperformed HOG, HOF, MBH and DT for action recognition.

\begin{table}[t]
\begin{center}
\caption{Impact of estimated body joints versus ground-truth body joints for JDD, P-CNN and two high-level pose features on subJHMDB dataset.}
\label{table:estimated_vs_groundtruth}
\begin{tabular}{|c|c|c|c|}
  \hline
  Method    & GT & Esti & Diff\\
  \hline
  JDD (conv5b) & \textbf{0.819} & \textbf{0.777} & \textbf{0.042}\\
  P-CNN \cite{/corr/CheronLS15/PCNN} & 0.725  & 0.668 & 0.057\\
  Pose \cite{Jhuang_2013_ICCV_subJHMDB_dataset} & 0.751 & 0.541 & 0.210 \\
  HLPF \cite{/corr/CheronLS15/PCNN} & 0.782 & 0.511 & 0.271 \\
  \hline
\end{tabular}
\end{center}
\end{table}

From Table \ref{table:estimated_vs_groundtruth}, we can see that JDD outperforms competing methods significantly on subJHMDB dataset.
The existing pose-based deep convolutional descriptors have not take full advantage of body joints.
The high-level pose features suffer from severe performance degradation when the body joint positions are inaccurate.
Note that we only use the JDD pooled from $conv5b$ and employ simple average pooling between clips. JDD achieves the best performance not only with ground-truth body joints, but also with estimated body joints, exceeding other methods in the order of 10\%. And the drop of accuracy for JDD is the smallest among all the descriptors which demonstrates that JDD is pretty robust to errors in pose estimation.

The superior performance of JDD compared with P-CNN which additionally used optical flow images as input demonstrates that we do not need to crop the images into multiple patches to advance accuracy as usual. The information in feature maps obtained by taking one image or video clip as input is abundant. We can take good advantage of it by keypoint pooling.

We further evaluate the robustness of JDD by adding white Gaussian noise to the ground-truth body joint positions in all the frames of subJHMDB dataset and Penn Action dataset. The noise has zero mean and the standard deviation $({\sigma}_x, {\sigma}_y)$ is proportional to the size of frame:

\begin{equation}
({\sigma}_x, {\sigma}_y) = \alpha \times (W,H)
\label{equation:sigma}
\end{equation}
where $\alpha$ is a coefficient indicating the ratio of noise intensity to the frame size.

\begin{figure}[t]
  \centering
  \subfigure[subJHMDB dataset]{
  \includegraphics[width=0.22\textwidth]{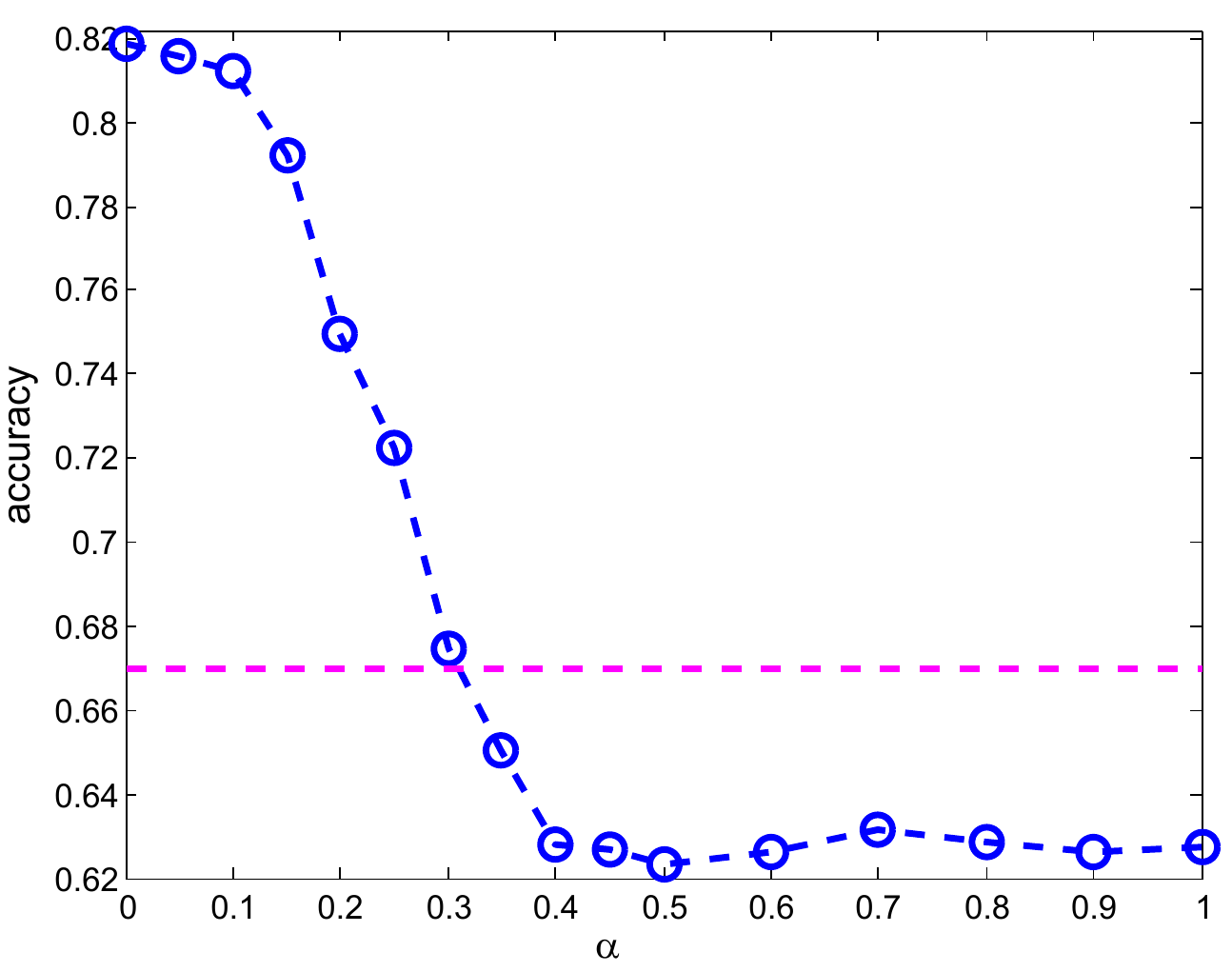}
  }
  \subfigure[Penn Action dataset]{
  \includegraphics[width=0.22\textwidth]{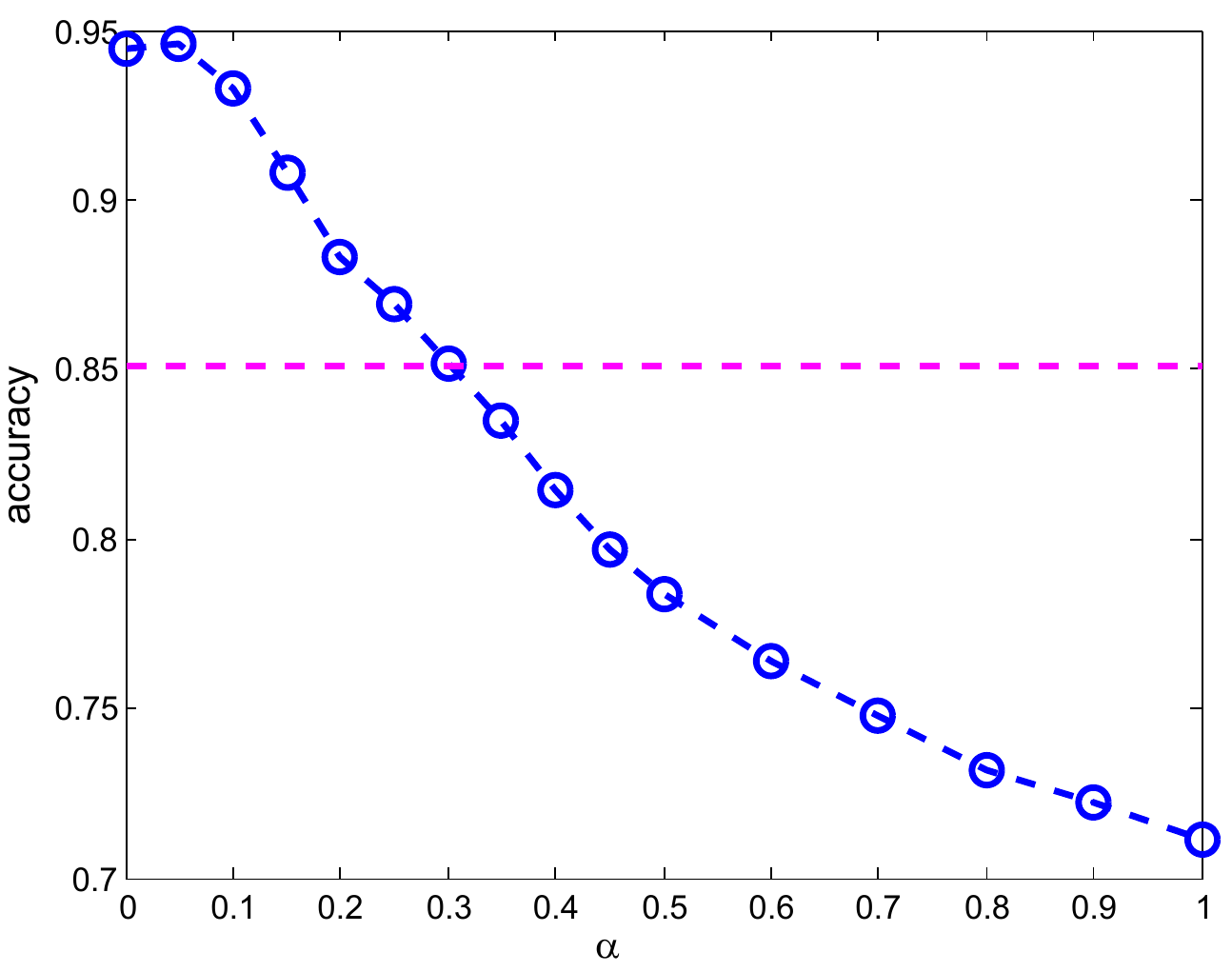}
  }
  \caption{The sensitivity of JDD from $conv5b$ to the noise added to the body joint positions. The coefficient $\alpha$ of x axis indicates the ratio of noise intensity to frame size. The magenta dashed line plots the accuracy of C3D $conv5b$ as a baseline.}
  \label{fig:addWGN}
\end{figure}

We plot the accuracies of JDD from $conv5b$ under different degrees of noise in Figure \ref{fig:addWGN}.
The magenta dashed line represents the accuracy of C3D $conv5b$ which concatenates all the activations on $conv5b$ without pooling.
The accuracy of JDD $conv5b$ is higher than C3D $con5b$ until the coefficient $\alpha$ is bigger than 0.3.
Take the video frames with $320 \times 240$px for example, a coefficient with 0.3 means that the standard deviation of noise is $(96,72)$ in pixel, which is a considerably large noise added to body joint positions.
The experimental results demonstrate that JDD is effective in extracting discriminative features while robust to noise.

\subsection{Analysis of two-stream bilinear C3D and other pooling models}   \label{section:attention_model_experiment}

\begin{table*}[t]
\begin{center}
\caption{Recognition accuracy of C3D \cite{C3D_2}, variations of cross-convolutional-layer pooling \cite{cross-layer-pooling} \cite{bilinear_CNN}, JDD, bilinear pooling with 3D attention model and jointly finetuned two-stream bilinear C3D on Penn Action dataset.* means the advanced version of aggregation by using max+min pooling in temporal dimension.}
\label{table:attention_Penn}
\begin{tabular}{|c|c|c|c|c|c|c|c|c|c|c|}
  \hline
   \multicolumn{2}{|c|}{Penn Action} & \tabincell{c}{C3D \\conv5b} & \tabincell{c}{C3D \\fc6 \cite{C3D_2}} & \tabincell{c}{conv4b$\times$\\conv5b} & \tabincell{c}{conv5a$\times$\\conv5b \cite{cross-layer-pooling}} & \tabincell{c}{conv5b$\times$\\conv5b \cite{bilinear_CNN}} & \tabincell{c}{JDD conv5b\\estimated joints} & \tabincell{c}{JDD conv5b\\ground truth} & \tabincell{c}{Heat map$\times$\\conv5b} & \tabincell{c}{Heat map$\times$\\conv5b *}
  \\
  \hline
  \hline
  \multicolumn{2}{|c|}{dimension} & 50176 & 4096 & 262144 & 262144 & 262144 & 122880 & 106496 & 106496 & 26624 \\
  \hline
  \hline
  \multirow{2}{*}{\tabincell{c}{if finetuned\\deep model}}
  & No  & 0.851 & 0.860 & 0.881 & 0.888 & 0.878 & 0.874 & \textbf{0.943} & 0.917 & \textbf{0.951}\\
  \cline{2-11}
  & Yes & 0.895 & 0.897 & 0.915 & 0.919 & 0.910 & 0.908 & \textbf{0.958} &0.926 & \textbf{0.953} \\

  \hline
\end{tabular}
\end{center}
\end{table*}

\begin{table*}[t]
\begin{center}
\caption{Recognition accuracy of C3D \cite{C3D_2}, variations of cross-convolutional-layer pooling \cite{cross-layer-pooling} \cite{bilinear_CNN}, JDD and bilinear pooling with transferred attention model on subJHMDB dataset. * means the advanced version of aggregation.}
\label{table:attention_transfer_subJHMDB}
\begin{tabular}{|c|c|c|c|c|c|c|c|c|}
  \hline
   subJHMDB & \tabincell{c}{C3D \\conv5b} & \tabincell{c}{C3D \\fc6 \cite{C3D_2}} & \tabincell{c}{conv5a$\times$\\conv5b \cite{cross-layer-pooling}} & \tabincell{c}{conv5b$\times$\\conv5b \cite{bilinear_CNN}} & \tabincell{c}{JDD conv5b\\estimated joints} & \tabincell{c}{JDD conv5b\\ground truth} & \tabincell{c}{Heat map$\times$\\conv5b} & \tabincell{c}{Heat map$\times$\\conv5b *}
  \\
  \hline
  \hline
  dimension & 50176 & 4096 & 262144 & 262144 & 122880 & 122880 & 106496 & 26624\\
  \hline
  \hline
  accuracy (not finetuned) & 0.670 & 0.689 & 0.742 & 0.775 & 0.777 & \textbf{0.819} & 0.788 & \textbf{0.797}\\
  \hline
\end{tabular}
\end{center}
\end{table*}

\begin{table*}[t]
\begin{center}
\caption{Recognition accuracy of C3D \cite{C3D_2}, the spatial net of two-stream CNN \cite{SimonyanZ2014_NIPS_TwoStream}, TDD \cite{/cvpr/Wang2015_TDD}, and bilinear pooling with transferred attention model on UCF101 dataset. We compare between the models using RGB images as input for fairness. * means the advanced version of aggregation.}
\label{table:attention_transfer_UCF101}
\begin{tabular}{|c|c|c|c|c|c|c|c|c|}
  \hline
  UCF101 & \tabincell{c}{C3D \\conv5b} & \tabincell{c}{C3D \\fc6 \cite{C3D_2}} & Spatial net \cite{SimonyanZ2014_NIPS_TwoStream} & \tabincell{c}{TDD FV \cite{/cvpr/Wang2015_TDD} \\Spatial conv4} & \tabincell{c}{TDD FV \cite{/cvpr/Wang2015_TDD} \\Spatial conv5} & \tabincell{c}{TDD FV \cite{/cvpr/Wang2015_TDD} \\Spatial conv4+conv5} & \tabincell{c}{Heat map$\times$\\conv5b} & \tabincell{c}{Heat map$\times$\\conv5b *}
  \\
  \hline
  \hline
  dimension & 50176 & 4096 & 4096 & 32768 & 32768 & 65536 & 106496 & 26624\\
  \hline
  \hline
  if finetuned & no & no & yes & yes & yes & yes & no & no\\
  \hline
  \hline
  accuracy   & 0.811 & 0.815 & 0.730 & 0.819 & 0.809 & 0.828 & 0.827 & \textbf{0.847} \\
  \hline
\end{tabular}
\end{center}
\end{table*}

In our proposed two-stream bilinear C3D model, we pool the spatio-temporal convolutional feature maps with the 3D heat maps predicted by the attention stream which is pre-trained with the guidance of body joint positions. We make a comparison of the prosed model with existing pooling models in this section.
The results of C3D original features, variations of cross-convolutional-layer pooling, JDD and two-stream bilinear pooling on Penn Action dataset are listed in Table \ref{table:attention_Penn}.
The symbol ``$\times$'' means bilinear product.

Cross-convolutional-layer pooling was proposed by \cite{cross-layer-pooling} for fine-grained image classification. It pooled one convolutional layer with the guidance of the successive layer.
The computation of pooling could be realized by bilinear product as described in Section \ref{section:JDD_with_bilinear_product}.
The idea of cross-convolutional-layer pooling can apply to any two convolutional layers.
If the last convolutional layer is multiplied by itself, it is a special case of the bilinear CNN model \cite{bilinear_CNN} with the configuration of using two identical network architectures.
We extend cross-convolutional-layer pooling to any two 3D convolutional layers as baselines.

For JDD, we use the algorithm in \cite{/cvpr/YangR11/subJHMDB/estimated_joints} to estimate the positions of body joints in Penn Action dataset without finetuning.
The number of body joints defined by the annotation of Penn Action dataset is 13, which is not equal to 15 defined with the skeleton estimation algorithm.
The latter contains two more body joints which are ``neck'' and ``belly'' than the former.
This is why the dimension of JDD with the estimated body joints is not equal to that with the ground-truth body joints as listed in Table \ref{table:attention_Penn}.
The per joint L1 distance error between the ground-truth body joint positions and the corresponding estimated positions is (68.52, 40.67) pixels in width and height.
The average ratio of L1 distance error to frame size is (0.145, 0.131).

For two-stream bilinear pooling, we pre-train the 3D attention stream as introduced in Section \ref{section:3D_attention_model}.
The ground-truth positions of keypoints in heat maps is computed with Coordinate Mapping for its good performance.
When training the attention model, we use the parameters of C3D convolutional layers as initialization. The training is stopped after 10000 iterations with mini-batches of 10 and learning rate of $10^{-7}$.
For finetuning the two-stream bilinear C3D with class label, the mini-batch is set to 5. We firstly using the learning rate of $10^{-4}$ to finetune the fully-connected layers with 5000 iterations, then we finetune the whole network with learning rate of $10^{-6}$. The learning rate is divided by 10 three times respectively after 10000, 20000, 20000 iterations.

As shown in Table \ref{table:attention_Penn}, the highest accuracy is obtained by JDD based on ground-truth body joints.
The result of JDD with estimated joints could be further improved with more precise predictions.
Excluding JDD based on annotated joints, pooling with heat maps regressed by 3D attention model achieves the best classification result.
With jointly finetuning the whole network of two-stream bilinear C3D supervised by class label, the performance is further improved.
By replacing the operations of intra-clip concatenation and inter-clip average pooling both with max+min pooling in temporal dimension for feature aggregation, the accuracy of the jointly finetuned two-stream bilinear C3D is increased by 3\%.

For subJHMDB which is too small to train a deep network and UCF101 which is not annotated with ground-truth body joints, the 3D attention model trained on Penn Action dataset after 10000 iterations with 90/10 training/testing split and $10^{-6}$ learning rate is transferred to them.
The recognition results of pooling models on subJHMDB and Penn Action dataset are listed in Table \ref{table:attention_transfer_subJHMDB} and Table \ref{table:attention_transfer_UCF101} respectively.
We do not enumerate the results of P-CNN which also belongs to pooling methods in this section since they have already been analysed in Section \ref{section:robustness_analysis}.
The feature dimension of P-CNN is 163840 which is higher than that of JDD and two-stream bilinear C3D.
The estimated body joints of subJHMDB dataset predicted with algorithm \cite{/cvpr/YangR11/subJHMDB/estimated_joints} is provided by the dataset.
For large-scale dataset UCF101, the computation cost of skeleton estimation algorithm as analysed in Section \ref{section:implementation_details} is too high (approximately 10 seconds for 1 frame). With the 3D architecture we use, it is able to process at 313 fps using a single K40 Tesla GPU.
It is unsuitable to apply cross-convolutional-layer pooling and the original bilinear CNN \cite{bilinear_CNN} to large-size dataset such as UCF101, because the feature dimensions of these models are high. In our proposed two-stream bilinear C3D, with the pre-trained 3D attention model which is guided with body joints, the number of channels of one stream is decreased, therefore the dimension of the pooled descriptor is reduced. With the advanced version of aggregation as introduced in Section \ref{Section:aggregation_methods}, the dimension is further reduced and the performance is improved.
On UCF101, we compare our methods with other models using RGB images as input. Although advanced performance can be achieved by taking optical flow images as input, the computational cost of optical flow is heavy for large-scale dataset. And not only competing methods, but also C3D can use optical flow images as input. Therefore, it is more fair to compare between methods that use only RGB images as input.
The trajectory-pooled deep-convolutional descriptor (TDD) \cite{/cvpr/Wang2015_TDD} pooled on the normalized 2D convolutional feature maps based on the trajectories of dense points.
In fact, the process of commutating dense trajectories include the computation of optical flow. TDD used the information of optical flow even in the spatial CNN.
TDD FV used Fisher vector to encode local TDDs over the whole video into a global super vector, where local TDDs were the descriptors of video clips obtained by aggregating the pooled feature vectors belonging to one trajectory with sum pooling.

Note that we do not finetune our proposed model on subJHMDB and UCF101. The results listed in Table \ref{table:attention_transfer_subJHMDB} and Table \ref{table:attention_transfer_UCF101} demonstrates that bilinear pooling with 3D attention model is generic and effective in extracting discriminative features in 3D convolutional layers. Our proposed approach outperforms competing methods with higher accuracy and lower feature dimension.

\subsection{Comparison with the state of the art}

\subsubsection{\textbf{Evaluation on subJHMDB Dataset}}

We compare our method with the state-of-the-art approaches on subJHMDB. As represented in Table \ref{table:subJHMDB}, our proposed JDD and two-stream bilinear C3D outperform competing methods significantly.

The methods classified to \textbf{Video features} use the raw video frames to extract features. They do not use ground-truth body joints, neither in training phrase nor in testing phrase.
The methods classified to \textbf{Pose features} use the ground-truth body joints for training. Some of these work also reported their results with annotated body joints in testing.
There are attempts to combine video features with pose features for performance improvement which are classified to \textbf{Pose+Video features}.

Among video features, C3D performs much better than Dense Trajectories (DT) and Improved Dense Trajectories encoded with Fisher vector (IDT-FV).
We increase the recognition of C3D with body joint guided pooling by about 10\% with estimated joins and 15\% with ground-truth body joints.
We have already made a brief comparison of JDD with pose features HLPF, Pose, and posed-based convolutional descriptor P-CNN in Section \ref{section:robustness_analysis}.
ACPS \cite{arXiv2016_PoseforAction_ActionforPose} is an action conditioned pictorial structure model for pose estimation that incorporated priors over actions.
Directly combining the video features and the pose features together does not necessarily lead to performance improvement, taking Pose+DT for example.
MST-AOG \cite{DBLP:conf/cvpr/WangNXWZ14} is a multiview spatio-temporal AND-OR graph representation for cross-view action recognition. Body joints were utilized to mine the discriminative parts.
ST-AOG \cite{/cvpr/NieXZ15_And-Or_Graph_subJHMDB_Penn} is also a spatio-temporal AND-OR graph which decomposed actions into poses, spatio-temporal parts, and parts.

Compared to other methods, JDD has a superior performance.
Fusing JDDs of $conv5b$ and $conv4b$ together further improves the performance benefiting from the complementarity between convolutional layers.
When ground-truth body joints are used, two-stream bilinear C3D is equivalent to JDD of $conv5b$. Therefore, they have the same accuracy in these configurations.
Different from JDD and P-CNN which are dependent on skeleton estimation algorithms to predict body joint positions in practical situation, two-stream bilinear C3D learns keypoints in feature maps automatically with the transferred 3D attention model.
The best accuracy without ground-truth body joints in testing is achieved by two-stream bilinear C3D with the advanced version of aggregation by using max+min pooling in temporal dimension, outperforming competing methods significantly.

It should be noted that, JDD and our proposed two-stream bilinear model fuse the information of body joints and 3D convolutional features in model level. They also can be combined with IDT-FV and other features in feature level or decision level as ACPS+IDT-FV did.

\begin{table}
\begin{center}
\caption{Recognition accuracy of the state-of-the-art approaches and our proposed models on subJHMDB dataset. * means the advanced version of aggregation.}
\label{table:subJHMDB}
\begin{tabular}{|c|c|c|c|}
  \hline
  \multirow{2}{*}{ }
  & \multirow{2}{*}{Method}
  & \multicolumn{2}{c|}{Accuracy}  \\
  \cline{3-4}
   &  &  \tabincell{c}{testing without \\GT joints}  &  \tabincell{c}{testing with \\GT joints} \\
  \hline
  \hline
  \multirow{3}{*}{\tabincell{c}{Video \\features}}
   & DT \cite{Jhuang_2013_ICCV_subJHMDB_dataset}            & 0.460  & - \\
   & IDT-FV \cite{wang:ICCV2013_iDT}                           & 0.609  & - \\
   & C3D fc6 \cite{C3D_2}                                      & 0.688  & - \\
  \hline
  \hline
  \multirow{3}{*}{\tabincell{c}{Pose \\features}}
   & HLPF \cite{/corr/CheronLS15/PCNN}                         & 0.511  & 0.782 \\
   & Pose \cite{Jhuang_2013_ICCV_subJHMDB_dataset}             & 0.541  & 0.751 \\
   & ACPS \cite{arXiv2016_PoseforAction_ActionforPose}         & 0.615  & -     \\
  \hline
  \hline
  \multirow{9}{*}{\tabincell{c}{Pose+\\Video \\features}}
   & MST-AOG \cite{DBLP:conf/cvpr/WangNXWZ14}                      & 0.453  & -     \\
   & Pose+DT \cite{Jhuang_2013_ICCV_subJHMDB_dataset}       & 0.529  & -     \\
   & ST-AOG \cite{/cvpr/NieXZ15_And-Or_Graph_subJHMDB_Penn}       & 0.612  & -     \\
   & P-CNN \cite{/corr/CheronLS15/PCNN}                        & 0.668  & 0.725 \\
   & ACPS+IDT-FV \cite{arXiv2016_PoseforAction_ActionforPose}  & 0.746  & -     \\
   & JDD (conv5b)                                              & 0.777  & 0.819 \\
   & JDD (conv5b+conv4b)                                       & 0.778  & 0.833 \\
   & JDD (conv5b+conv4b) *                                     & 0.772  & \textbf{0.837} \\
   & two-stream bilinear C3D                                    & 0.788  & 0.819 \\
   & two-stream bilinear C3D *                                  & \textbf{0.797}  & 0.826 \\
  \hline
\end{tabular}
\end{center}
\end{table}

\subsubsection{\textbf{Evaluation on Penn Action Dataset}}   \label{section:Evaluation_on_Penn}

\begin{table}
\begin{center}
\caption{Recognition accuracy of the state-of-the-art approaches and our proposed models on Penn Action dataset. * means the advanced version of aggregation.}
\label{table:Penn}
\begin{tabular}{|c|c|c|c|}
  \hline
  \multirow{2}{*}{ }
  & \multirow{2}{*}{Method}
  & \multicolumn{2}{c|}{Accuracy}  \\
  \cline{3-4}
   &  &  \tabincell{c}{testing without \\GT joints}  &  \tabincell{c}{testing with \\GT joitns} \\
  \hline
  \hline
  \multirow{4}{*}{\tabincell{c}{Video \\features}}
   & DT \cite{DBLP:journals/ijcv/WangKSL13}                 & 0.734  & - \\
   & STIP  \cite{iccv/ZhangZD13_Penn_Action_dataset}           & 0.829  & - \\
   & IDT-FV \cite{wang:ICCV2013_iDT}                           & 0.920  & - \\
   & C3D fc6 \cite{C3D_2}                                      & 0.860  & - \\
  \hline
  \hline
  \multirow{3}{*}{\tabincell{c}{Pose \\features}}
   & ACPS \cite{arXiv2016_PoseforAction_ActionforPose}         & 0.790  & -     \\
   & Actemes \cite{iccv/ZhangZD13_Penn_Action_dataset}         & 0.794  & -     \\
   & Action Bank \cite{/cvpr2012/action_bank}     & 0.839  & -     \\

  \hline
  \hline
  \multirow{7}{*}{\tabincell{c}{Pose+\\Video \\features}}
   & MST-AOG \cite{DBLP:conf/cvpr/WangNXWZ14}                      & 0.740  & -     \\
   & ST-AOG \cite{/cvpr/NieXZ15_And-Or_Graph_subJHMDB_Penn}       & 0.855  & -     \\
   & ACPS+IDT-FV \cite{arXiv2016_PoseforAction_ActionforPose}  & 0.929  & -     \\
   & P-CNN \cite{/corr/CheronLS15/PCNN}                        & \textbf{0.953}  & 0.977  \\
   & JDD (conv5b)                                              & 0.874  & 0.943  \\
   & JDD(conv5b+conv4b)                                        & 0.893  & 0.957 \\
   & JDD(conv5b+conv4b) *                                      & 0.938  & \textbf{0.981} \\
   & two-stream bilinear C3D                                   & 0.926  & 0.943      \\
   & \tabincell{c}{two-stream bilinear C3D *}          & \textbf{0.953} & 0.971      \\
  \hline
\end{tabular}
\end{center}
\end{table}


\begin{figure}[t]
  \centering
  \subfigure[C3D fc6]{
  \includegraphics[width=0.22\textwidth]{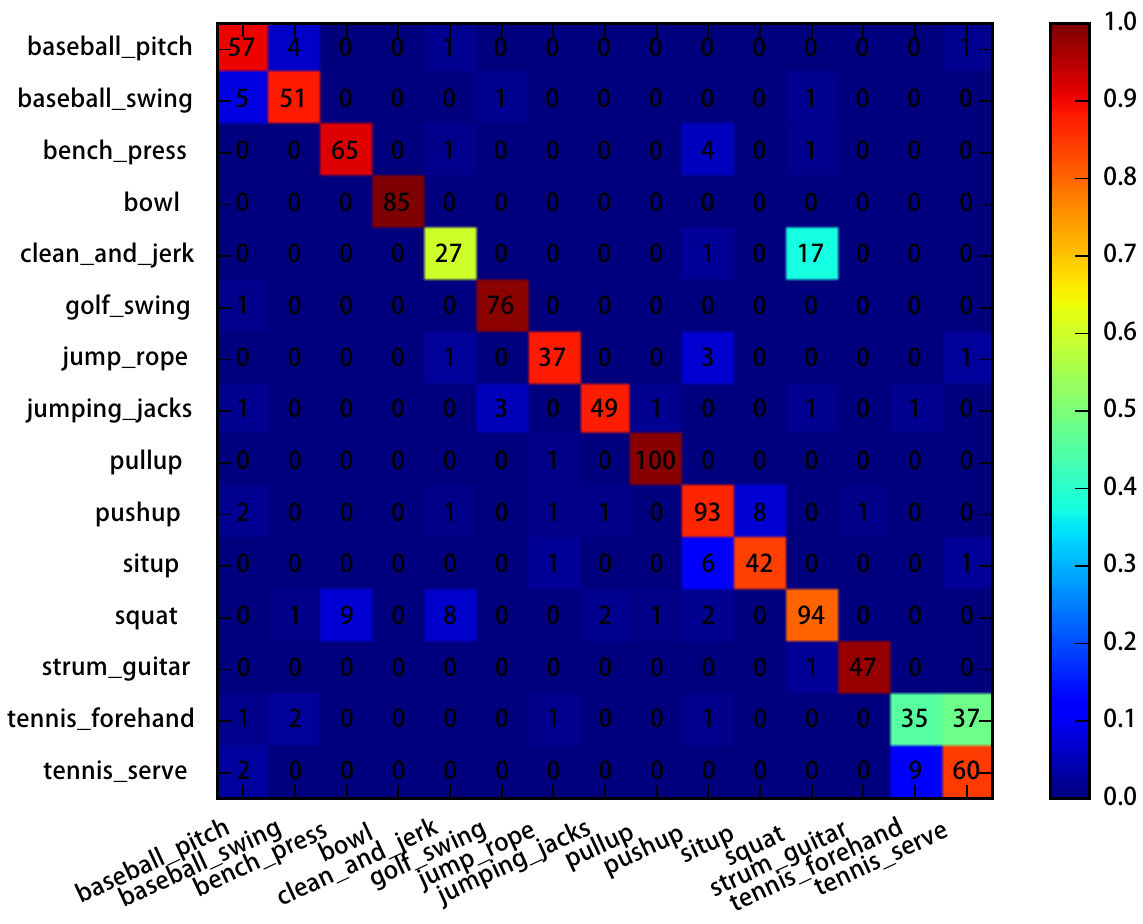}
  }
  \subfigure[two-stream bilinear C3D]{
  \includegraphics[width=0.22\textwidth]{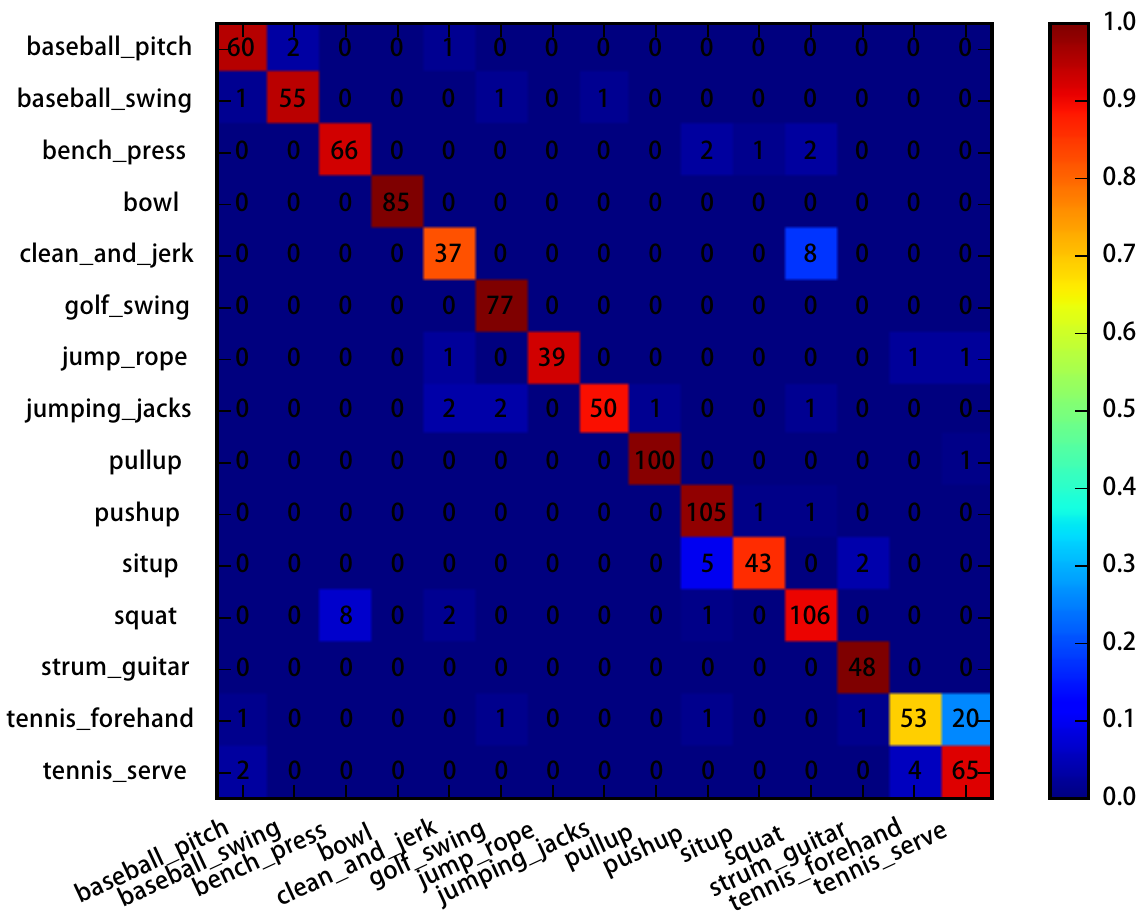}
  }
  \caption{The confusion matrixes obtained by C3D $fc6$ and two-stream bilinear C3D on Penn Action dataset.}
  \label{fig:ConfusionMatrix}
\end{figure}

%

There are videos with invisible body joints on Penn Action dataset. Take the category of ``playing guitar'' for example, less than one third of a person is visible.
There also exist errors in annotations.
Nevertheless, compared with other methods, our proposed JDD and two-stream bilinear C3D still have superior performance as shown in Table \ref{table:Penn}.
Except the methods which have been introduced before,
Actemes \cite{iccv/ZhangZD13_Penn_Action_dataset} used body joint positions to discover patches that were clustered in the spatio-temporal keypoint configuration space.
Action bank \cite{/cvpr2012/action_bank} represented actions on the basis of holistic spatio-temporal templates.

The performance of C3D is inferior to IDT-FV on this dataset. The reason of C3D's inferiority is probably that C3D resized the video images with too much information loss. The widths of frames in Penn Action dataset vary from 270 to 482 pixels, while the heights of frames vary from 204 to 480 pixels. Most frame resolutions are $480 \times 270$ and $480 \times 360$ which are larger than $320 \times 240$ in subJHMDB dataset. By resizing frames to $171 \times 128$ and cropping to $112 \times 112$ patches as input, C3D changed the ratio and resolution of video frames in Penn Action dataset too much.
We follow the experimental setting of C3D to resize and crop frames. While P-CNN took multiple $224 \times 224$ image patches as input.
We think the performances of C3D-based methods are damaged because of small input images. By using larger images as input, the accuracy of C3D, JDD and two-stream bilinear C3D should be improved.

In addition,
IDT is a hand-crafted feature based on optical flow tracking and histograms of low-level gradients. IDT-FV encoded IDT with fisher vector.
P-CNN benefits from optical flow in another way by taking optical flow images as an extra input.
However it is time-consuming to compute IDT-FV and optical flow. In \cite{C3D_2}, there has been runtime analysis which shows that C3D is 91 times faster than IDT \cite{wang:ICCV2013_iDT} and 274 times faster than Brox's optical flow methods \cite{Brox_optical_flow}.
Our proposed two-stream bilinear C3D is even free from the sophisticated skeleton estimation algorithms.
If only for performance improvement, it is feasible to combine C3D features, JDD and two-stream bilinear features, which are high-level semantic descriptors, with IDT-FV and optical flow since they are complementary to each other.

Note that the authors of \cite{/cvpr/NieXZ15_And-Or_Graph_subJHMDB_Penn} removed the action ``playing guitar'' and several other videos because less than one third of a person is visible in those data. While we do not remove any videos. This illustrates that JDD is robust for occlusion. The authors \cite{/cvpr/NieXZ15_And-Or_Graph_subJHMDB_Penn} also corrected the errors of annotated body joints which remain at the left-top corner of image by training a regression model to predict their positions. While we do nothing to rectify the positions of body joints. We even do not finetune the skeleton estimation algorithm on Penn Action dataset. This indicates that JDD is not sensitive to error.


We visualize the confusion matrixes obtained by C3D $fc6$ and two-stream bilinear C3D in Figure
\ref{fig:ConfusionMatrix}.
The most confusing categories of C3D are ``clean and jerk'' versus ``squat'', ``tennis forehand'' versus ``tennis serve'' due to their similarity in appearance and motion. With the ability of extracting discriminative spatio-temporal features in convolutional feature maps, two-stream bilinear C3D performs much better than C3D $fc6$ features.
If we use the advanced version of aggregation instead of the basic version, the accuracy of two-stream bilinear C3D could be further improved to 95.3\% and the accuracy of fusing JDDs from $conv5b$ and $conv4b$ could be 98.1\% with ground-truth body joints.


The result in Table \ref{table:Penn} is consistent with that of Table \ref{table:subJHMDB}.
The best accuracy with ground-truth joints is obtained by fusing JDDs of multiple layers together.
The best accuracy without annotation in testing is obtained with two-stream bilinear C3D.
JDDs are verified to make the best of body joints and 3D CNN to form discriminative video descriptors.
Furthermore, two-stream bilinear C3D unified the computations of JDD in an end-to-end framework which is independent of skeleton estimation algorithms.

\subsubsection{\textbf{Evaluation on UCF101 Dataset}}

\begin{table}
\begin{center}
\caption{Recognition accuracy of the state-of-the-art approaches and our proposed models on UCF101 dataset. * means the advanced version of aggregation.}
\label{table:UCF101}
\begin{tabular}{|c|c|c|}
  \hline
  Method & Accuracy \\
  \hline
  IDT \cite{wang:ICCV2013_iDT}      & 0.762     \\
  IDT-FV  \cite{wang:ICCV2013_iDT}  & \textbf{0.859}    \\

  \hline
  ImageNet \cite{JiaSDKLGGD14_Caffe}  & 0.688  \\
  CNN-M-2048 \cite{SimonyanZ2014_NIPS_TwoStream} & 0.730  \\
  VGGNet \cite{WangXW2015_arXiv_Very_Deep_TwoStream_ConvNets} & 0.784 \\

  \hline
  LRCN \cite{/cvpr2015/LRCN}    & 0.711  \\
  LSTM composite model \cite{/icml2015/LSTM} & 0.758  \\

  \hline
  Slow Fusion network \cite{cvpr/KarpathyTSLSF14_larege_sacle_video_classification} & 0.654  \\
  C3D fc6 \cite{C3D_2} & 0.815  \\
  two-stream bilinear C3D   & 0.827  \\
  two-stream bilinear C3D * & 0.847  \\
  \hline
\end{tabular}
\end{center}
\end{table}

We compare our proposed two-stream bilinear C3D against hand-crafted features, deep convolutional networks, convolutional networks combined with recurrent neural networks and other 3D CNN models on UCF101 which is a challenging dataset with occlusions, large-scale variations and complex scenes. 

The results are shown in Table \ref{table:UCF101}.
These models use RGB video frames as input.
IDT-FV improves the performance of IDT significantly with Fisher vector encoding. However, as analysed in Section \ref{section:Evaluation_on_Penn}, dense trajectory based methods are computationally intensive and have a high time cost.
ImageNet \cite{JiaSDKLGGD14_Caffe} refers to the deep features extracted using Caffe's ImageNet pre-trained model.
Very deep two-stream convolutional network \cite{WangXW2015_arXiv_Very_Deep_TwoStream_ConvNets} replaced the CNN-M-2048 architecture \cite{DBLP:conf/bmvc/ChatfieldSVZ14} used in two-stream convolutional network \cite{SimonyanZ2014_NIPS_TwoStream} with VGGNet \cite{VGGNet} which has smaller convolutional kernel sizes, smaller convolutional strides, and deeper network architectures. The results of CNN-M-2048 and VGGNet are obtained with well designed training.
LRCN \cite{/cvpr2015/LRCN} and LSTM composite model \cite{/icml2015/LSTM} are two CNN-RNN based methods. They used long short term memory (LSTM) units to model the temporal evolutions over video frames which are represented by convolutional features \cite{JiaSDKLGGD14_Caffe,VGGNet}. The convolutional network and the RNN network can be trained jointly.
Slow fusion network \cite{cvpr/KarpathyTSLSF14_larege_sacle_video_classification} used 3D convolutions and average pooling in its first 3 convolutional layers.

Leaving out the results with the aid of optical flow and dense trajectories, C3D performs pretty well among competing methods. We report the result of C3D $fc6$ reproduced by us.
With the transferred attention stream and bilinear pooling, our proposed two-stream bilinear C3D outperforms all the other RGB-based models significantly even without finetuning.

\section{Conclusions}

In this paper, we propose a novel joints-pooled 3D deep convolutional descriptor (JDD) which can take advantages of body joints and C3D for action recognition.
The positions of body joints are used to sample discriminative points from feature maps generated by C3D.
Furthermore, we propose an end-to-end trainable two-stream bilinear C3D model which formulates the body joint guided feature pooling as a bilinear product operation. The two-stream bilinear C3D learns keypoints in 3D feature maps, captures the spatio-temporal features and pools activations in a unified framework.
Experiments show that
the best recognition result with ground-truth body joints is obtained by multi-layer fusion of JDDs. Without precise annotated body joints, two-stream bilinear C3D achieves the highest performance among competing methods.
Promising experimental results in real-world datasets demonstrate the effectiveness and robustness of JDD and two-stream bilinear C3D in video-based action recognition.

In the future, we will consider to integrate optical flow and dense trajectories into our proposed two-stream bilinear model in an efficient way. Also, temporal models such as RNN will be tried to replace the aggregation methods used now. Meanwhile, the problem of overfitting when training the CNN-RNN network which is doubly deep in spatial and temporal dimensions with limited data need be solved with network designment and data augmentation.



\bibliographystyle{IEEEtran}
\small
\bibliography{IEEEabrv,sigproc}

\end{document}